\documentclass[sigconf]{acmart}
\AtBeginDocument{%
  }

\copyrightyear{2026}
\acmYear{2026}
\setcopyright{cc}
\setcctype{by}
\acmConference[MM '26]{Proceedings of the 34th ACM International Conference on Multimedia}{November 10--14, 2026}{Rio de Janeiro, Brazil}
\acmBooktitle{Proceedings of the 34th ACM International Conference on Multimedia (MM '26), November 10--14, 2026, Rio de Janeiro, Brazil}
\acmDOI{10.1145/3767308.3836021}
\acmISBN{979-8-4007-2213-4/2026/11}

\usepackage{algorithm}
\usepackage{algorithmic}
\newcommand{\rfdtablefont}{\fontsize{8.6pt}{9.9pt}\selectfont}
\begin{document}

\title{R4DGS: Referring Segmentation in 4D Gaussian Splatting}

\author{Bangpu Chen}
\authornote{Both authors contributed equally to this research.}
\affiliation{%
  \institution{China University of Geosciences}
  \city{Wuhan}
  \country{China}
}

\author{Yaxuan Li}
\authornotemark[1]

\affiliation{%
  \institution{Xinjiang University}
  \city{Urumqi}
  \country{China}
}

\author{Shirui Peng}
\affiliation{%
  \institution{University of Nottingham Ningbo China}
  \city{Ningbo}
  \country{China}
}

\author{Xiangtian Si}
\affiliation{%
  \institution{China University of Geosciences}
  \city{Wuhan}
  \country{China}
}

\author{Liuxin Chu}
\affiliation{%
  \institution{China University of Geosciences}
  \city{Wuhan}
  \country{China}
}

\author{Xitong Cao}
\affiliation{%
  \institution{China University of Geosciences}
  \city{Wuhan}
  \country{China}
}

\author{Hongbo Jin}
\affiliation{%
  \institution{Peking University}
  \city{Shenzhen}
  \country{China}
}

\author{Jiayu Ding}
\correspondingauthor
\affiliation{%
  \institution{Peking University}
  \city{Shenzhen}
  \country{China}
}
% \email{jyding25@stu.pku.edu.cn}

\renewcommand{\shortauthors}{Bangpu Chen et al.}

\begin{abstract}
Dynamic representations based on 4D Gaussian Splatting (4DGS) have shown strong capabilities in reconstruction and rendering, yet current methods remain limited in supporting complex natural-language referring in dynamic scenes. Existing 4D Gaussian methods are primarily designed for rendering-oriented dynamic modeling, while current semantic extensions often rely on scene-specific semantic optimization or are evaluated on relatively simple query settings. In this paper, we study Referring 4D Gaussian Splatting (R4DGS), a task for grounding temporally varying, multi-target or reasoning-intensive, and zero-target or distractor expressions in dynamic 4D scenes. To support this task, we introduce R4D-Bench-QA, a benchmark with structured query annotations. We further present ReferGaussian, a unified framework built upon 4DGS that converts low-level dynamic Gaussians into persistent, language-addressable entities for training-free referring inference. ReferGaussian organizes mask-supported Gaussian entities in an EntityBank, uses Multimodal Large Language Models (MLLMs) to decompose complex queries into trackable object phrases and structured referring constraints, and performs constraint-aware grounding without scene-specific semantic optimization. Experiments on R4D-Bench-QA and the 4D LangSplat protocol demonstrate improved temporal target selection and competitive spatiotemporal grounding without scene-specific semantic training. Project resources are available at \url{https://chenbangpu.github.io/R4DGS/}.
\end{abstract}




\begin{teaserfigure}
  \centering
  \includegraphics[width=\textwidth]{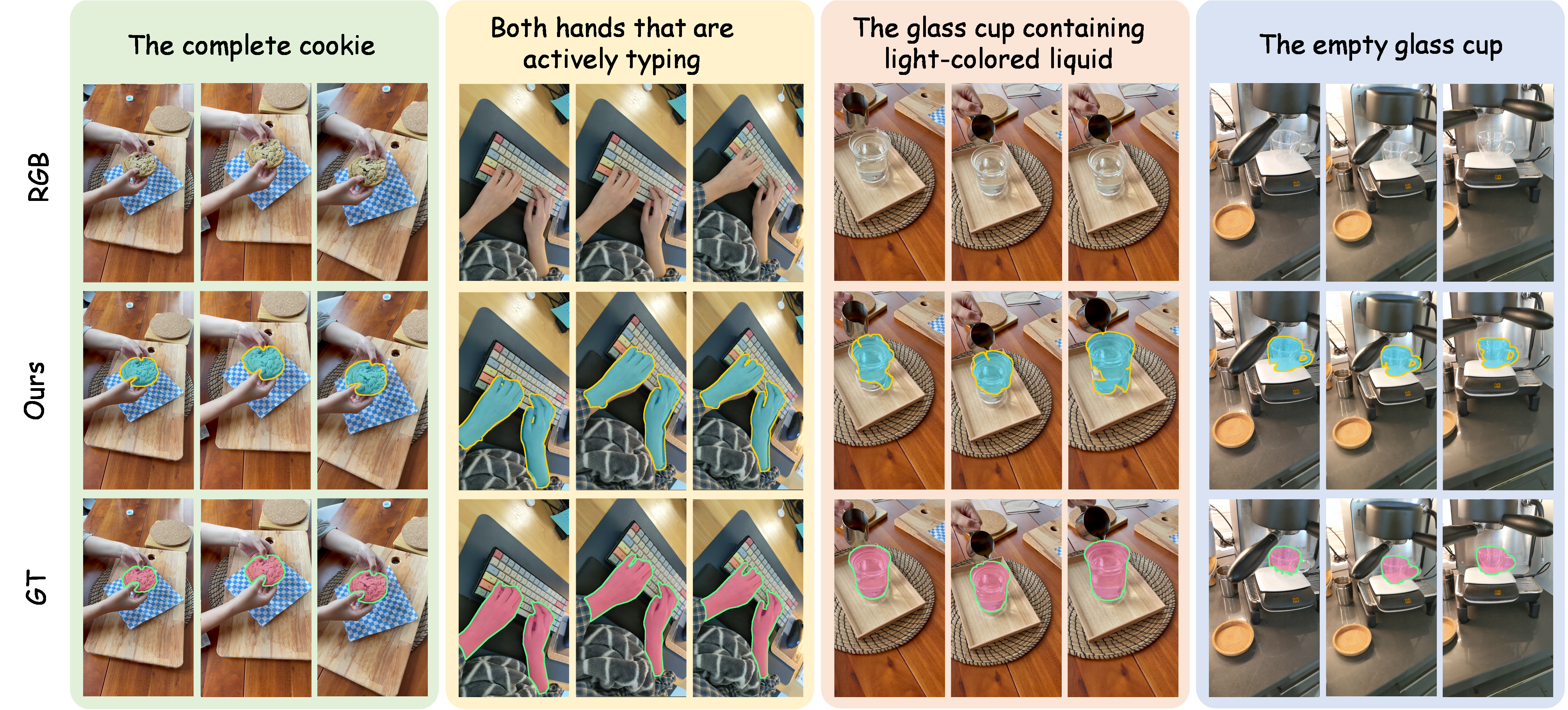}
  \caption{Example visualizations of ReferGaussian for dynamic and multi-target referring in 4D scenes.}
  \Description{Qualitative referring segmentation results for a complete cookie, two actively typing hands, a glass cup containing light-colored liquid, and an empty glass cup across three frames.}
  \label{fig:teaser}
\end{teaserfigure}

\maketitle

\section{Introduction}
3D Gaussian Splatting (3DGS) has become a powerful representation for static scene reconstruction and novel-view rendering, offering high visual fidelity and efficient rasterization \cite{kerbl20233dgs}. Building on this foundation, recent dynamic Gaussian methods extend Gaussian primitives across time through deformation, temporal conditioning, or explicit spacetime modeling, substantially advancing dynamic scene reconstruction and free-viewpoint rendering \cite{luiten2024dynamic3dgs,yang2024deformable3dgs,li2024sgfs,wu20244dgs,yang20244dgsiclr}. Among these developments, 4D Gaussian Splatting (4DGS) has emerged as an effective representation for dynamic scene modeling, enabling high-fidelity rendering with strong efficiency \cite{wu20244dgs,yang20244dgsiclr}. Together, these advances establish Gaussian splatting as a practical substrate for dynamic visual modeling.

At the same time, recent work has begun to study language grounding and semantic reasoning on Gaussian representations. In static 3D scenes, methods such as LangSplat \cite{qin2024langsplat}, Gaussian Grouping \cite{ye_gaussian_2024}, OpenGaussian \cite{wu2024opengaussian}, Dr. Splat \cite{jun-seong_dr_2025}, Segment then Splat \cite{lu2025segmentthensplat}, and ReferSplat \cite{he2025refersplat} connect Gaussian primitives with language or segmentation supervision. In dynamic settings, SA4D \cite{ji2024sa4d} and 4D LangSplat \cite{li20254dlangsplat} extend Gaussian-based semantics to time-varying scenes. However, many existing methods are optimized for category-level querying, scene-specific semantic fitting, or mask prediction, and are not designed for expressions involving temporal state changes, interaction cues, multi-target reasoning, or deliberately absent targets.

\begin{figure}[!t]
  \centering
  \includegraphics[width=\linewidth,trim=80 0 80 0,clip]{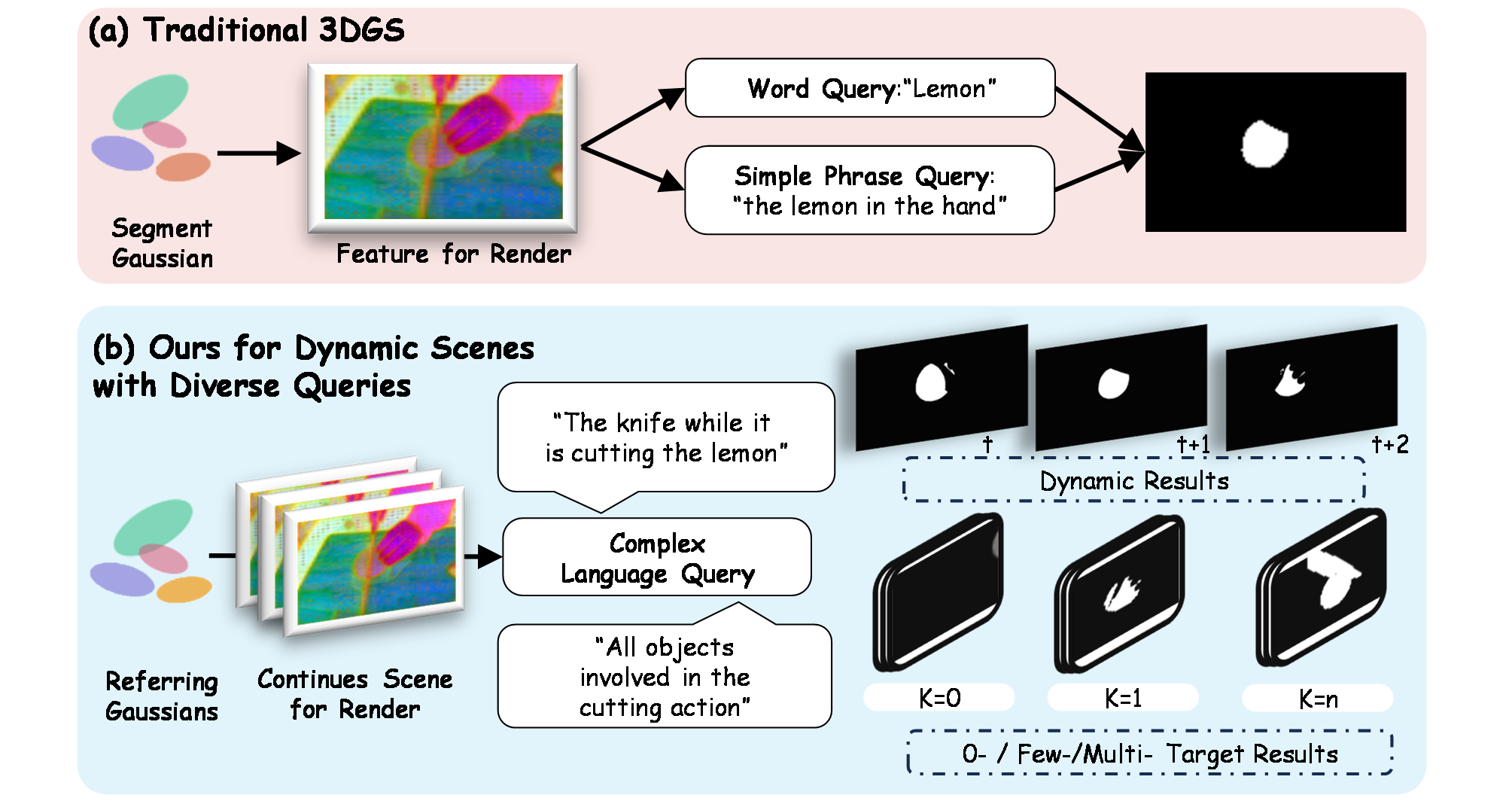}
  \caption{Task comparison between Gaussian semantic queries and R4DGS. R4DGS resolves temporal, relational, exclusion, and cardinality constraints into zero-, one-, or multi-target mask sequences.}
  \Description{Task-definition comparison figure for R4DGS, including zero-target, few-target, and multi-target outcomes in panel (d).}
  \label{fig:task-definition}
\end{figure}

This gap becomes critical when the query involves temporal states, relations, or set-level constraints. A practical model should handle expressions such as \emph{the cookie broken into smaller pieces}, \emph{the complete lemon before the cut}, \emph{all objects except the hand and the metal cup}, or descriptions with no valid referent. These queries require temporal reasoning, entity disambiguation, set-level constraints, and uncertainty-aware rejection. Figure~\ref{fig:task-definition} summarizes the transition from conventional semantic querying to dynamic referring.

Figure~\ref{fig:teaser} provides example visualizations of ReferGaussian for dynamic and multi-target referring in 4D scenes.

\begin{figure*}[!t]
  \begin{minipage}[t]{\columnwidth}
    \centering
    \includegraphics[width=\linewidth]{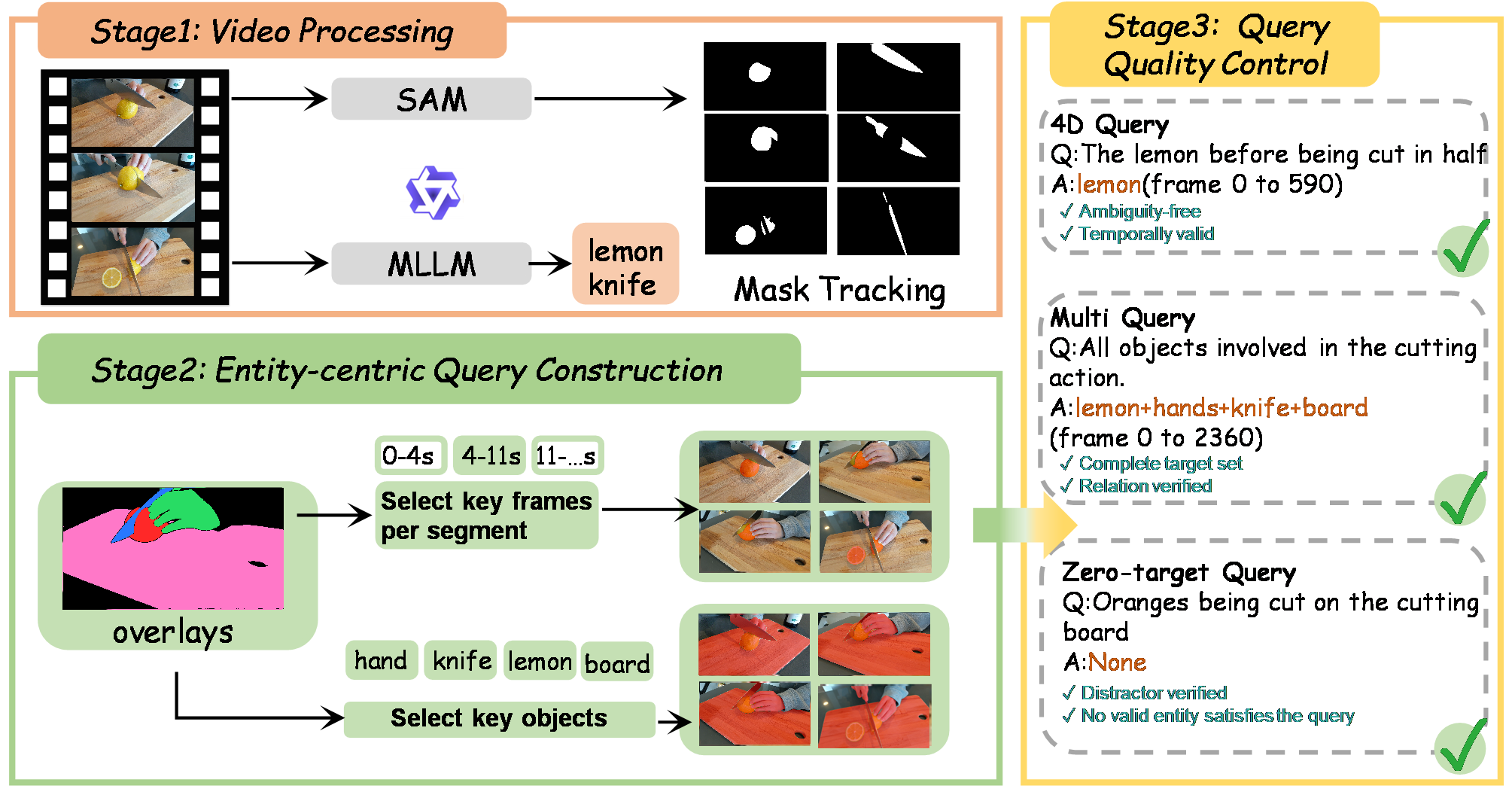}
    \captionof{figure}{Construction pipeline of R4D-Bench-QA for track extraction, entity-centric annotation, and quality control. These stages convert source videos into trackable entities and verified entity-centric query annotations.}
    \Description{Three-stage construction pipeline for R4D-Bench-QA, comprising video processing, entity-centric query construction, and query quality control.}
    \label{fig:dataset-construction}
  \end{minipage}\hfill
  \begin{minipage}[t]{\columnwidth}
    \centering
    \includegraphics[width=\linewidth]{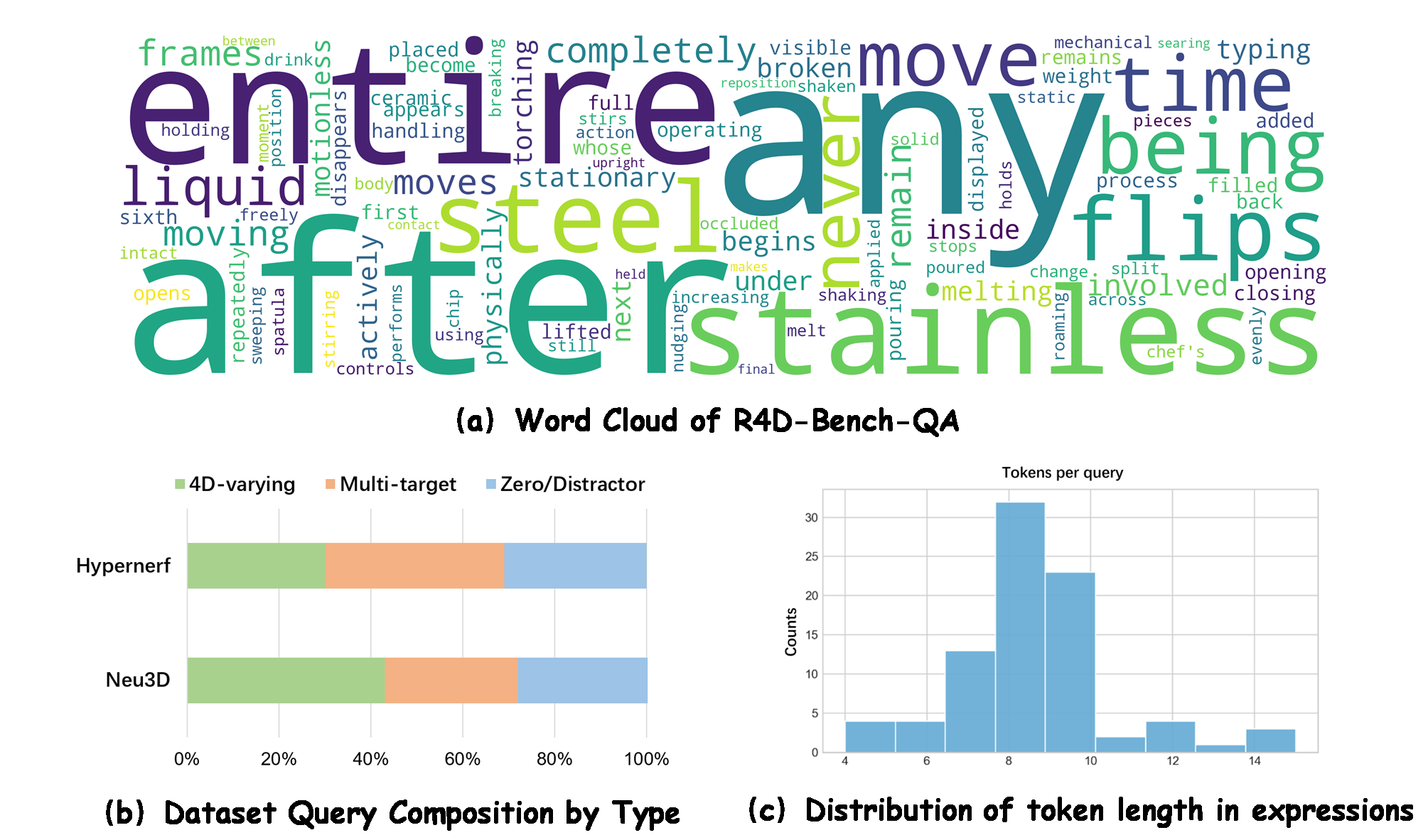}
    \captionof{figure}{Statistics of R4D-Bench-QA across vocabulary, query types, and expression length. The bars summarize three query categories and two scene sources.}
    \Description{Statistics of R4D-Bench-QA.}
    \label{fig:benchmark}
  \end{minipage}
\end{figure*}

To address these limitations, we define Referring 4D Gaussian Splatting (R4DGS), which requires a 4D Gaussian system to resolve complex natural-language referring expressions in dynamic scenes and produce temporally grounded, set-valued segmentation masks. Recognizing the lack of a benchmark for this task, we introduce R4D-Bench-QA (QA denotes Query Annotations), which evaluates three challenging categories: 4D-varying queries, multi-target or reasoning queries, and zero-target or distractor queries.

These observations reveal an abstraction gap between low-level Gaussian primitives and the entity-level units referred to by natural language. Dynamic Gaussian backbones provide efficient time-conditioned scene representations, but their primitives are not directly organized into persistent, language-addressable entities. Meanwhile, SA4D and 4D LangSplat enrich dynamic scenes with segmentation identities or learned language fields \cite{ji2024sa4d,li20254dlangsplat}, yet remain primarily oriented toward segmentation-centric outputs or scene-specific semantic fitting. Consequently, temporal-state composition, multi-entity selection, and zero-target rejection are not naturally supported within a unified referring protocol.

Motivated by this gap, we propose ReferGaussian, a training-free referring framework built upon 4DGS. Starting from a reconstructed 4D Gaussian scene, it uses mask-supported multi-view evidence to form persistent Gaussian entities, records their appearance, motion, relations, and valid temporal support in an EntityBank, and resolves structured referring constraints over this bank. MLLMs serve as the \emph{Refer-Planner} for query decomposition and constraint extraction, while Grounded-SAM supplies 2D mask evidence. The final prediction is produced through explicit and inspectable entity-level selection, without scene-specific semantic training.

Our contributions are summarized as follows:
\begin{enumerate}
  \item \looseness=-1 We formulate Referring 4D Gaussian Splatting (R4DGS), a novel task for language-grounded dynamic scene understanding involving temporal state changes, multi-target reasoning, and distractor rejection.
  \item \looseness=-1 We establish R4D-Bench-QA, the first benchmark for R4DGS, featuring structured query annotations and pixel-level spatiotemporal supervision to address the current lack of evaluation standards.
  \item We propose ReferGaussian, a training-free framework that organizes low-level dynamic Gaussians into persistent entities addressable through language and performs constraint-aware 4D referring over an EntityBank.
\end{enumerate}

\section{Related Work}

\noindent\textbf{Dynamic Scene Reconstruction.}
NeRF \cite{mildenhall2020nerf} and its dynamic variants \cite{Park_2021_ICCV,pumarola2021dnerf,li2021HyperNerf,wang2021nsff,Cao_2023_CVPR} laid the foundation for neural dynamic view synthesis. 3D Gaussian Splatting \cite{kerbl20233dgs} replaced volumetric rendering with explicit anisotropic Gaussian primitives, making real-time rendering practical. Recent work extends Gaussian splatting to dynamic scenes through deformation, canonicalization, or explicit spacetime modeling, including Dynamic 3D Gaussians \cite{luiten2024dynamic3dgs}, Deformable 3D Gaussians \cite{yang2024deformable3dgs}, 4D Gaussian Splatting \cite{wu20244dgs}, Real-time Photorealistic Dynamic Scene Representation and Rendering with 4D Gaussian Splatting \cite{yang20244dgsiclr}, and Spacetime Gaussian Feature Splatting \cite{li2024sgfs}. These methods deliver strong rendering quality, yet their objective remains view synthesis. Our target is language-conditioned reasoning over dynamic entities. This requires temporal support that remains stable across rendering, downstream grouping, candidate ranking, and set-level decisions under ambiguous natural language. In practice, this difference is critical when multiple visually similar candidates coexist but only one satisfies temporal or relational constraints across a full event window.

\noindent\textbf{Language-Grounded Semantic Reasoning.}
Visual semantic reasoning has advanced across various domains. In 2D video understanding, several works explore semantic tasks such as open-vocabulary and referring segmentation \cite{wu2022language, wu2023onlinerefer, tang2023temporal, seo2020urvos, luo2023soc, jin2026vistamitigatingsemanticinertia, jin2026tirflowactivevideosearch, jin2026himachierarchicalmacromicrolearning}.
For static 3D scenes, current research explores two primary directions: open-vocabulary understanding and segmentation \cite{kerr2023lerf, peng2023openscene, qin2024langsplat, ye_gaussian_2024, wu2024opengaussian, jun-seong_dr_2025, ding2026extrin, ding20263dinstructionambiguitydetection, zhang2025cobgs, lu2025segmentthensplat, piekenbrinck2025opensplat3d}, and referring expression grounding represented by ReferSplat \cite{he2025refersplat}.
In dynamic contexts, while SA4D \cite{ji2024sa4d} and 4D LangSplat \cite{li20254dlangsplat} enable time-sensitive language querying, they primarily focus on semantic segmentation, with limited coverage of complex referring understanding. To our knowledge, referring grounding in dynamic Gaussian scenes has not been systematically formulated or evaluated.
Unlike existing methods that rely on scene-specific semantic retraining or learned language fields, ReferGaussian resolves referents through entity-level matching within a query-reusable structure.
This training-free design supports robust reasoning over temporal states and set-valued targets through persistent entity records and explicit constraints.

\noindent\textbf{Foundation Models.}
Large vision-language and segmentation foundation models provide strong building blocks for language-grounded scene reasoning. Qwen3-VL \cite{bai2025qwen3} supports structured interpretation of natural-language expressions, while Grounding DINO \cite{liu2023groundingdino}, Segment Anything \cite{kirillov2023sam}, and Grounded-SAM \cite{ren2024groundedsam} enable text-conditioned mask extraction. Recent multimodal systems \cite{liu2024improved,wang2024cogvlm,chen2024internvl} further improve prompt following and visual grounding.
ReferGaussian uses these models as modular components for query parsing and visual evidence extraction within a structured entity-centric pipeline. This decomposition improves diagnostic ability: query decomposition errors, mask evidence errors, and entity association errors can be analyzed separately.

\begin{table*}[t]
  \caption{Comparison of representative Gaussian-language benchmarks, domains, and query support. Checkmarks denote supported capabilities.}
  \label{tab:benchmark-comparison}
  \centering
  \rfdtablefont
  \setlength{\tabcolsep}{3.1pt}
  \renewcommand{\arraystretch}{1.24}
  \begin{tabular*}{\textwidth}{@{\extracolsep{\fill}}llccllccccc@{}}
    \toprule
    Dataset & Source & Scenes & Query Scale & Domain & Task Type & Temporal & \begin{tabular}[c]{@{}c@{}}Sentence-level\\Referring\end{tabular} & \begin{tabular}[c]{@{}c@{}}Dynamic\\Semantics\end{tabular} & \begin{tabular}[c]{@{}c@{}}Temp.\\Query\end{tabular} & \begin{tabular}[c]{@{}c@{}}Spa.\\Query\end{tabular} \\
    \midrule
    LERF & LangSplat & 4 & 208 words & 2D & Open-vocab seg. & $\times$ & $\times$ & $\times$ & $\times$ & $\checkmark$ \\
    Ref-LERF & ReferSplat & 4 & 63 inst. & 2D & Ref. seg. & $\times$ & $\times$ & $\times$ & $\times$ & $\checkmark$ \\
    ScanNet & OpenGaussian & 10 & 84 words & 3D & Open-vocab seg. & $\times$ & $\times$ & $\times$ & $\times$ & $\times$ \\
    \begin{tabular}[c]{@{}l@{}}lerf\_ovs\_\\langsplat\end{tabular} & \begin{tabular}[c]{@{}l@{}}Segment then\\Splat\end{tabular} & 5 & 208 words & 3D & Open-vocab seg. & $\times$ & $\times$ & $\times$ & $\times$ & $\checkmark$ \\
    \begin{tabular}[c]{@{}l@{}}HyperNeRF-\\Annotation\end{tabular} & 4D LangSplat & 4 & 13 sentences & 4D & Open-vocab seg. & $\checkmark$ & $\times$ & $\checkmark$ & $\checkmark$ & $\checkmark$ \\
    \textbf{R4D-Bench (Ours)} & \textbf{R4DGS} & \textbf{12} & \textbf{89 sentences} & \textbf{4D} & \textbf{Ref. seg.} & \textbf{$\checkmark$} & \textbf{$\checkmark$} & \textbf{$\checkmark$} & \textbf{$\checkmark$} & \textbf{$\checkmark$} \\
    \bottomrule
  \end{tabular*}
\end{table*}

\section{Task and Benchmark}

In this section, we first formalize Referring 4D Gaussian Splatting (R4DGS) and clarify its task setting. We then introduce R4D-Bench-QA, our benchmark for dynamic 4D referring understanding, together with its construction pipeline, annotation protocol, query taxonomy, and evaluation metrics.

\subsection{Referring 4D Gaussian Splatting (R4DGS)}

Given a multi-view dynamic video, a precomputed 4D Gaussian scene representation, and a natural-language query $q$, R4DGS identifies the referred entities and produces their frame-aligned segmentation masks. The referent may be a single entity, multiple entities, or an empty set, depending on the spatiotemporal, relational, and set-level constraints expressed by the query.

R4DGS adds three requirements to conventional referring: (1) \emph{Temporally varying queries} specify an event stage or a state before or after a transition. (2) \emph{Set-valued and reasoning-intensive queries} express conjunction, exclusion, quantification, or relations among entities. (3) \emph{Zero-target and distractor queries} may describe no valid referent and therefore return an empty prediction.

\subsection{R4D-Bench-QA}

To evaluate R4DGS systematically, we construct R4D-Bench-QA, a query-annotation benchmark with deliberate coverage of three query categories and pixel-level spatiotemporal annotations. Its 12 scenes come from two dynamic-scene sources: six HyperNeRF~\cite{li2021HyperNerf} scenes provide 47 queries, and six Neu3D~\cite{li2022neural3dvideo} scenes provide 42 queries, for 89 densely annotated sentence-level queries in total. Table~\ref{tab:benchmark-comparison} compares R4D-Bench-QA with representative datasets used by Gaussian-language methods. Existing resources are either limited to static 2D/3D scenes or primarily designed for open-vocabulary segmentation, with limited support for sentence-level dynamic referring. R4D-Bench-QA jointly supports dynamic scenes, sentence-level referring, time-varying semantics, temporal constraints, and spatial grounding.

\noindent\textbf{Benchmark Construction.}
As illustrated in Figure~\ref{fig:dataset-construction}, the benchmark is constructed in three stages. In Stage~1, each dynamic video is processed to obtain temporally associated mask tracks and a vocabulary of trackable object candidates. Segmentation models provide frame-level mask evidence, while an MLLM assists in identifying concise object phrases that remain visually traceable throughout the sequence.

In Stage~2, annotators inspect the mask overlays and divide each sequence into semantically meaningful temporal segments according to object-state changes, interactions, and event boundaries. Key frames are selected from each segment to summarize its visual content. Based on the tracked objects and temporal segments, annotators construct entity-centric referring expressions and associate each expression with its complete target set, valid temporal support, and frame-aligned masks. Multi-target annotations include the complete set of entities satisfying the query.

Stage~3 performs query-specific quality control. Temporally varying queries are checked for unambiguous state descriptions and valid event boundaries. Multi-target queries are checked for relational consistency and completeness of the annotated target set. Zero-target queries are retained only when no annotated entity satisfies all query constraints. Three expert annotators perform verification, while MLLM-based checks flag ambiguous or inconsistent cases for human inspection.

\begin{figure*}[!t]
  \centering
  \includegraphics[width=\textwidth]{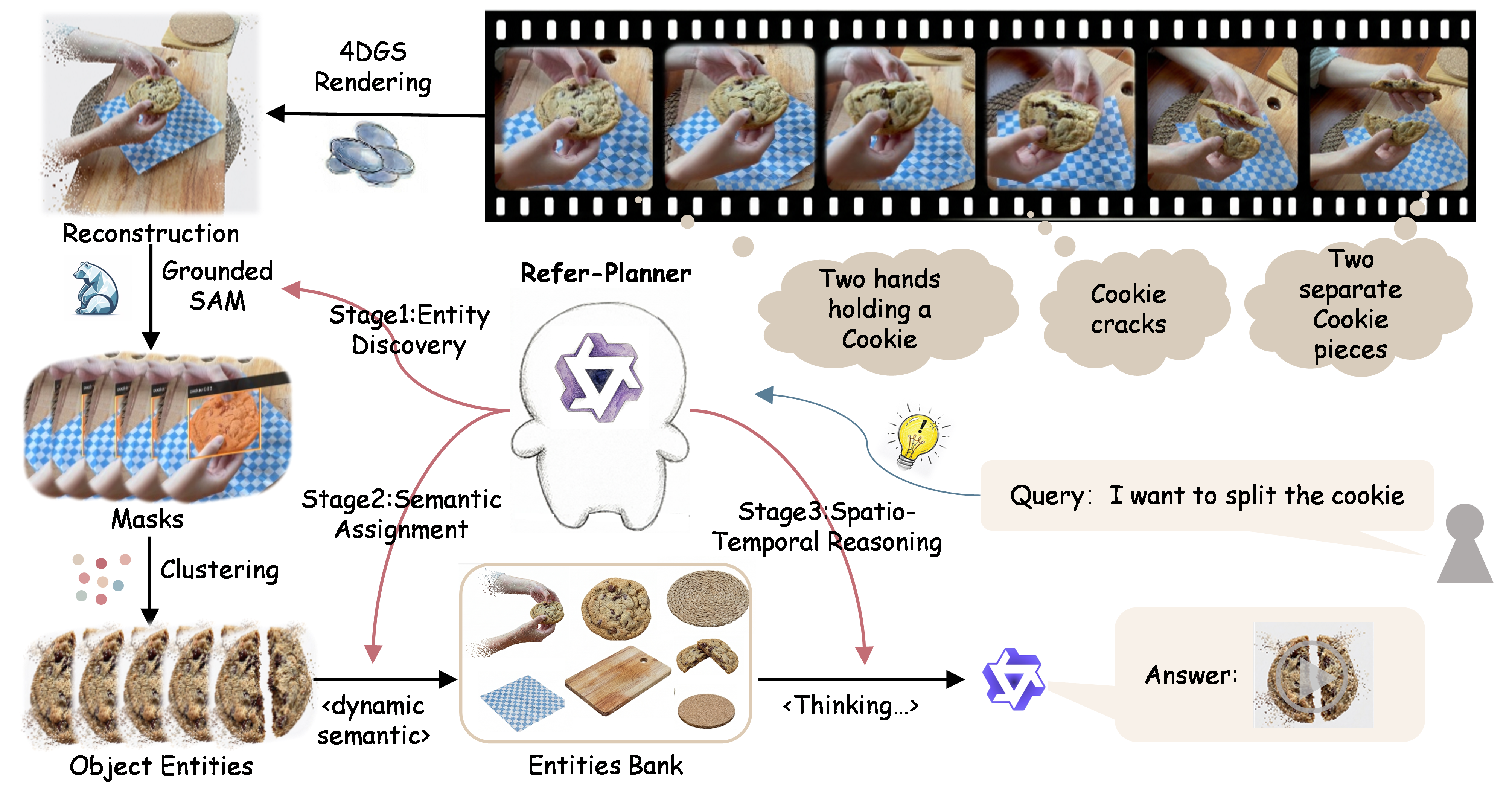}
  \caption{Overview of ReferGaussian, from Gaussian entity construction to constraint-aware spatiotemporal reasoning. The three stages discover persistent entities, organize reusable evidence, and resolve structured queries.}
  \Description{ReferGaussian framework with query-guided mask tracking, mask-supported Gaussian entity construction, EntityBank formation, and constraint-aware referent selection.}
  \label{fig:architecture}
\end{figure*}

\noindent\textbf{Query Construction and Taxonomy.}
R4D-Bench-QA covers three categories: 4D variation, multi-target reasoning, and zero-target or distractor cases. Together, they span state transitions, temporal ordering, conjunction, disjunction, exclusion, object interactions, and target-cardinality constraints. Query wording is varied through attributes, relational phrases, event descriptions, and referring structures while preserving the underlying target semantics. This controlled variation prevents evaluation from depending on a narrow set of prompt templates.

Figure~\ref{fig:benchmark} summarizes the benchmark vocabulary, query-category composition, and expression-length distribution. Together, these statistics show that R4D-Bench-QA contains both concise object references and longer compositional expressions involving temporal or set-level constraints.

\noindent\textbf{Annotation Representation.}
For each query $q$, the annotation contains the relevant frame set $\mathcal{F}_q$, the target entities, and a binary ground-truth mask $M^{*}_{q,t}$ for every evaluated frame $t \in \mathcal{F}_q$. For a multi-target query, $M^{*}_{q,t}$ is the frame-wise union of the masks of all valid target entities. A zero-target query has $M^{*}_{q,t}=\varnothing$ for all evaluated frames. Predictions are represented analogously by $\hat{M}_{q,t}$, obtained by rasterizing the selected Gaussian entities.

\subsection{Evaluation Metrics}

We adopt the time-sensitive protocol of 4D LangSplat \cite{li20254dlangsplat}. It evaluates temporal target activation and spatial mask quality, with Acc denoting frame-wise temporal activation accuracy. Let the ground-truth and predicted active-frame sets be $\mathcal{T}^{*}_q=\{t\in\mathcal{F}_q:|M^{*}_{q,t}|>0\}$ and $\hat{\mathcal{T}}_q=\{t\in\mathcal{F}_q:|\hat{M}_{q,t}|>0\}$.

\noindent\textbf{Temporal Accuracy.}
The temporal accuracy of query $q$ is
\begin{equation}
\operatorname{Acc}(q)
=\frac{1}{|\mathcal{F}_q|}
\sum_{t\in\mathcal{F}_q}
\mathbb{1}\!\left[
\mathbb{1}[t\in\hat{\mathcal{T}}_q]
=\mathbb{1}[t\in\mathcal{T}^{*}_q]
\right].
\label{eq:temporal-accuracy}
\end{equation}
This metric penalizes both missed valid intervals and false activations outside the annotated temporal support.

\noindent\textbf{Video IoU.}
Following the same protocol, video IoU is
\begin{equation}
\operatorname{vIoU}(q)=
\begin{cases}
\displaystyle
\frac{1}{|\hat{\mathcal{T}}_q\cup\mathcal{T}^{*}_q|}
\sum_{t\in\hat{\mathcal{T}}_q\cap\mathcal{T}^{*}_q}
\frac{|\hat{M}_{q,t}\cap M^{*}_{q,t}|}
{|\hat{M}_{q,t}\cup M^{*}_{q,t}|},
& |\hat{\mathcal{T}}_q\cup\mathcal{T}^{*}_q|>0,\\[8pt]
1, & |\hat{\mathcal{T}}_q\cup\mathcal{T}^{*}_q|=0.
\end{cases}
\label{eq:viou}
\end{equation}
Frames outside the temporal intersection contribute zero, so vIoU penalizes temporal errors, false targets, and inaccurate mask boundaries. For a valid zero-target query, vIoU is one only when the prediction remains empty. ZeroAcc is the percentage of zero-target queries with empty predictions throughout the evaluated frames. Incomplete or unresolved outputs count as failures.

The reported Acc and vIoU average their query-level scores:
\begin{equation}
\operatorname{Acc}=\frac{1}{|\mathcal{Q}|}
\sum_{q\in\mathcal{Q}}\operatorname{Acc}(q),
\quad
\operatorname{vIoU}=\frac{1}{|\mathcal{Q}|}
\sum_{q\in\mathcal{Q}}\operatorname{vIoU}(q).
\end{equation}
For multi-target queries, evaluation uses the frame-wise union of all selected and annotated target masks. 

We also report Acc and vIoU for each of the three query categories separately on HyperNeRF and Neu3D. These six source-category slices expose whether a model's behavior is consistent across both scene sources and all query types; Figure~\ref{fig:planner-sensitivity} visualizes this breakdown for the planner analysis.

\section{Method}

ReferGaussian addresses the abstraction gap between low-level dynamic Gaussian primitives and the entity-level units referred to by natural language. Given a precomputed 4D Gaussian scene, our method constructs persistent Gaussian entities from multi-frame visual evidence, organizes them into an EntityBank, and performs constraint-aware referent selection without optimizing an additional scene-specific semantic field.

\subsection{Preliminaries}

\noindent\textbf{3D Gaussian Splatting.}
3D Gaussian Splatting represents a scene with a collection of anisotropic Gaussians $g_i=(\boldsymbol{\mu}_i,\mathbf{r}_i,\mathbf{s}_i,\mathbf{c}_i,\alpha_i)$, where $\boldsymbol{\mu}_i$ is the center, $\mathbf{r}_i$ and $\mathbf{s}_i$ parameterize rotation and scale, and $\mathbf{c}_i$ and $\alpha_i$ denote color and opacity, respectively~\cite{kerbl20233dgs}. The covariance is $\boldsymbol{\Sigma}_i=\mathbf{R}_i\mathbf{S}_i\mathbf{S}_i^{\top}\mathbf{R}_i^{\top}$, where $\mathbf{R}_i$ and $\mathbf{S}_i$ are constructed from $\mathbf{r}_i$ and $\mathbf{s}_i$. After projection, the color at pixel $\mathbf{p}$ is alpha-composited in depth order:
\begin{equation}
\mathbf{C}(\mathbf{p})
=\sum_{i=1}^{N}T_i(\mathbf{p})\alpha_i(\mathbf{p})\mathbf{c}_i,
\qquad
T_i(\mathbf{p})=\prod_{j<i}\left(1-\alpha_j(\mathbf{p})\right).
\label{eq:3dgs-rendering}
\end{equation}

\noindent\textbf{4D Gaussian Splatting.}
Dynamic Gaussian methods condition Gaussian primitives on time. A common formulation predicts position, rotation, and scale offsets with a temporal deformation function $\mathcal{D}_{\theta}$~\cite{wu20244dgs,yang2024deformable3dgs}:
\begin{equation}
\begin{aligned}
\mathcal{D}_{\theta}(g_i,t)
&=(\Delta\boldsymbol{\mu}_i^t,\Delta\mathbf{r}_i^t,\Delta\mathbf{s}_i^t),\\
(\boldsymbol{\mu}_i^t,\mathbf{r}_i^t,\mathbf{s}_i^t)
&=(\boldsymbol{\mu}_i,\mathbf{r}_i,\mathbf{s}_i)
+\mathcal{D}_{\theta}(g_i,t).
\end{aligned}
\label{eq:4dgs-deformation}
\end{equation}
Equation~\eqref{eq:3dgs-rendering} then applies to the time-conditioned primitives. In this work, the reconstructed 4D Gaussian scene is provided as input, and our contribution begins with organizing its primitives for referring understanding.

\begin{table}[t]
  \caption{Quantitative comparison on R4D-Bench-QA. Acc is frame-wise temporal activation accuracy; vIoU combines temporal localization and mask overlap.}
  \label{tab:main-r4dbench}
  \centering
  \rfdtablefont
  \setlength{\tabcolsep}{3.5pt}
  \renewcommand{\arraystretch}{1.20}
  \begin{tabular*}{\columnwidth}{@{\extracolsep{\fill}}lcc@{}}
    \toprule
    Method & \multicolumn{2}{c}{R4D-Bench-QA} \\
    \cmidrule(lr){2-3}
    & Acc (\%) $\uparrow$ & vIoU (\%) $\uparrow$ \\
    \midrule
    Segment then Splat & 55.6 & 28.4 \\
    4D LangSplat & 58.4 & 32.1 \\
    \textbf{ReferGaussian (Ours)} & \textbf{76.5} & \textbf{34.4} \\
    \bottomrule
  \end{tabular*}
\end{table}

\begin{algorithm}[t]
\caption{Mask-supported Gaussian entity selection with six-frame iterative validation and render-back stopping.}
\label{alg:selection}
\fontsize{7.8pt}{8.8pt}\selectfont
\begin{algorithmic}[1]
\REQUIRE Gaussians $\mathcal{G}$; tracked masks $\{\mathbf{M}_t\}$; projections $\{\Pi_t\}$; active frames $\mathcal{T}^{+}$
\ENSURE selected Gaussian entity $\mathcal{G}_{\mathrm{sel}}$
\STATE initialize candidate $\mathcal{G}_{\mathrm{sel}}$, best candidate $\mathcal{G}_{\mathrm{best}}$, best IoU $b\leftarrow-1$, and stale count $c\leftarrow0$
\FOR{$r=1$ to $10$}
  \STATE sample $\mathcal{T}_r\subseteq\mathcal{T}^{+}$, where $|\mathcal{T}_r|\leq\min(|\mathcal{T}^{+}|,6)$
  \STATE update Gaussian support using Eq.~\eqref{eq:gaussian-support} and the accumulated evidence
  \STATE update $\mathcal{G}_{\mathrm{sel}}$ from the support ranking and render-back feedback
  \STATE rasterize $\mathcal{G}_{\mathrm{sel}}$ and compute mean mask IoU on $\mathcal{T}_r$
  \IF{$\mathrm{IoU}>b$}
    \STATE $b\leftarrow\mathrm{IoU}$; $\mathcal{G}_{\mathrm{best}}\leftarrow\mathcal{G}_{\mathrm{sel}}$; $c\leftarrow0$
  \ELSE
    \STATE $c\leftarrow c+1$
  \ENDIF
  \IF{$\mathrm{IoU}\geq0.70$ or $c\geq3$}
    \STATE \textbf{break}
  \ENDIF
\ENDFOR
\RETURN $\mathcal{G}_{\mathrm{best}}$
\end{algorithmic}
\end{algorithm}

\noindent\textbf{Language-Grounded Gaussian Fields.}
LangSplat attaches a language feature to each Gaussian and renders it with the alpha-compositing weights used for color~\cite{qin2024langsplat}:
\begin{equation}
\mathbf{F}(\mathbf{p},t)
=\sum_{i=1}^{N}T_i(\mathbf{p},t)\alpha_i(\mathbf{p},t)\mathbf{f}_i(t).
\label{eq:language-splatting}
\end{equation}
4D LangSplat extends language splatting to dynamic scenes with object-wise MLLM descriptions and time-dependent features for object status changes~\cite{li20254dlangsplat}. It produces a learned language field for open-vocabulary segmentation. R4DGS uses explicit referential units that support joint temporal, relational, cardinality, exclusion, and empty-target constraints.

\subsection{Framework}

Given dynamic Gaussians $\mathcal{G}$, calibrated video frames $\{\mathbf{I}_t\}$, and a query $q$, ReferGaussian follows three semantic stages.

\emph{Stage~1: query-guided entity discovery} uses an MLLM-based Refer-Planner to extract trackable subject phrases and coarse temporal cues. Grounded-SAM2 produces phrase-conditioned mask tracks, which are iteratively lifted from sampled frames into Gaussian-supported candidate entities.

\emph{Stage~2: EntityBank construction} summarizes each candidate using representative observations, active temporal segments, appearance and state descriptions, motion cues, and relations. These records form an inspectable EntityBank over the reconstructed scene.

\emph{Stage~3: constraint-aware referent selection} validates the structured query and selects candidate entities and temporal segments under identity, temporal, relational, cardinality, exclusion, and zero-target constraints. The selected Gaussian supports are finally rasterized to produce frame-aligned masks.

Figure~\ref{fig:architecture} illustrates the overall pipeline. This decomposition changes the reasoning unit from a pixel or frame-wise feature match to an entity with persistent Gaussian support. Stage~1 establishes spatial and temporal correspondence between 2D evidence and 4D primitives, Stage~2 makes the entities language-addressable, and Stage~3 produces a globally consistent set-valued answer.

\subsection{Mask-Supported Persistent Entity Construction}

For a subject phrase produced by the Refer-Planner, let $\mathbf{M}_t$ be its tracked mask and $\Pi_t(g_i)$ the projected footprint of Gaussian $g_i$. We aggregate mask agreement over sampled frames with one score:
\begin{equation}
S_i(\mathcal{T})=
\frac{\displaystyle\sum_{t\in\mathcal{T}}u_i^t
\frac{|\Pi_t(g_i)\cap\mathbf{M}_t|}{|\Pi_t(g_i)|+\epsilon}}
{\displaystyle\sum_{t\in\mathcal{T}}u_i^t+\epsilon}.
\label{eq:gaussian-support}
\end{equation}
Here, $u_i^t$ is a projection-validity indicator. It is active when the Gaussian has positive depth, its projected center lies inside the image, and its rasterized footprint contributes alpha mass. This definition follows the visibility checks used by the 4DGS renderer.

Algorithm~\ref{alg:selection} iteratively updates a support-ranked candidate and validates it by rendering its mask back to the sampled frames. Each round uses six active frames, or all active frames when fewer are available. We run at most ten rounds, stop when mean render-back IoU reaches $0.70$, and also stop after three validation rounds without improvement. The best validated candidate is retained.

\begin{table*}[!t]
  \caption{Quantitative comparison of time-sensitive querying on the 4D LangSplat HyperNeRF protocol, with per-scene and average Acc and vIoU.}
  \label{tab:4dlangsplat-protocol}
  \centering
  \rfdtablefont
  \setlength{\tabcolsep}{2.8pt}
  \renewcommand{\arraystretch}{1.10}
  \begin{tabular*}{\textwidth}{@{\extracolsep{\fill}}l*{10}{c}@{}}
    \toprule
    Method & \multicolumn{2}{c}{americano} & \multicolumn{2}{c}{chickchicken} & \multicolumn{2}{c}{split-cookie} & \multicolumn{2}{c}{espresso} & \multicolumn{2}{c}{Average} \\
    \cmidrule(lr){2-3}\cmidrule(lr){4-5}\cmidrule(lr){6-7}\cmidrule(lr){8-9}\cmidrule(lr){10-11}
    & Acc & vIoU & Acc & vIoU & Acc & vIoU & Acc & vIoU & Acc & vIoU \\
    \midrule
    LangSplat~\cite{qin2024langsplat} & 45.19 & 23.16 & 53.26 & 18.20 & 73.58 & 33.08 & 44.03 & 16.15 & 54.01 & 22.65 \\
    4D LangSplat~\cite{li20254dlangsplat} & 89.42 & 66.07 & \textbf{96.73} & \textbf{90.62} & \textbf{95.28} & \textbf{83.14} & 81.89 & 49.20 & 90.83 & \textbf{72.26} \\
    \textbf{ReferGaussian (Ours)} & \textbf{99.43} & \textbf{68.13} & 82.89 & 60.87 & 90.91 & 55.79 & \textbf{98.80} & \textbf{67.01} & \textbf{93.01} & 62.95 \\
    \bottomrule
  \end{tabular*}
\end{table*}

\begin{figure*}[!t]
  \centering
  \includegraphics[width=\textwidth]{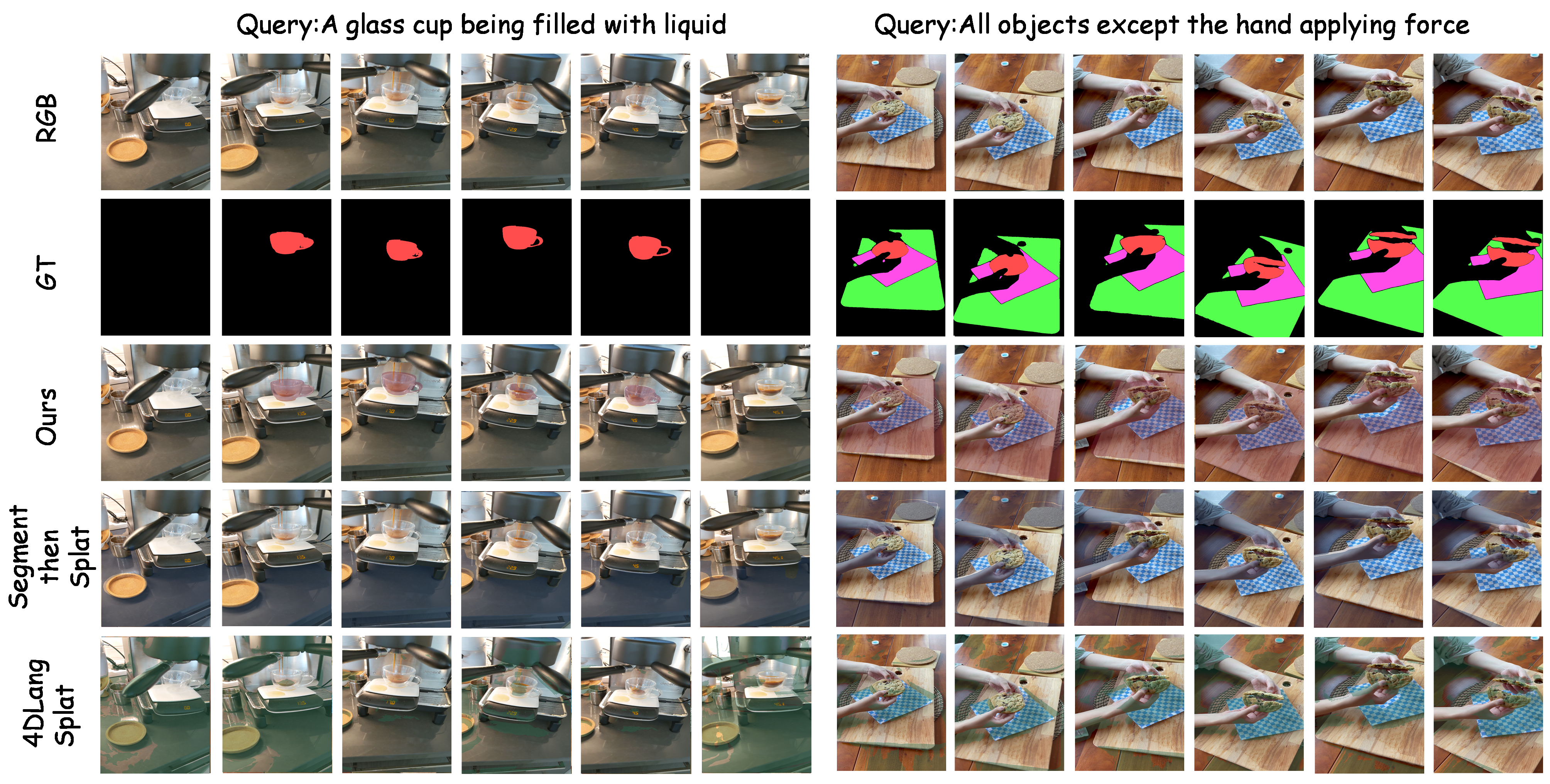}
  \caption{Qualitative comparison on R4D-Bench-QA for temporal-state and exclusion queries. Columns show query types, and rows compare frame-aligned predictions.}
  \Description{Representative qualitative comparison on R4D-Bench-QA with two query columns and method rows for RGB, ground truth, ReferGaussian, Segment then Splat, and 4D LangSplat.}
  \label{fig:qual-results}
\end{figure*}

The selected Gaussian set becomes a candidate entity $e_m$ with support $\mathcal{G}_m$. Its temporal support $\mathcal{T}_m$ is inherited from the active frames of the associated Stage~1 mask track, with adjacent active frames grouped into connected segments. The track identity is preserved during lifting. Appearance, motion, and temporal overlap connect recurring observations; incompatible tracks or trajectories remain separate. The resulting entity represents a physical candidate together with one or more temporally valid states.

\subsection{EntityBank Construction and Constraint-Aware Querying}

In Stage~2, each candidate is stored as
\begin{equation}
e_m
=\bigl(
\mathrm{id}_m,
\mathcal{G}_m,
\mathcal{T}_m,
\mathcal{X}_m,
\mathcal{A}_m,
\mathcal{R}_m
\bigr),
\label{eq:entity-record}
\end{equation}
where $\mathcal{G}_m$ is its Gaussian support, $\mathcal{T}_m$ contains candidate active segments, $\mathcal{X}_m$ contains representative crops and track summaries, and $\mathcal{A}_m$ and $\mathcal{R}_m$ contain appearance or state attributes and inter-entity relations. The collection $\mathcal{B}=\{e_m\}_{m=1}^{M}$ forms EntityBank. Constructed entity records and visual evidence are cached and reused when subsequent queries involve the same tracked objects.

The MLLM-based Refer-Planner converts each query into a schema containing trackable subject phrases, identity and state attributes, relational and exclusion constraints, temporal predicates, optional target cardinality, and an explicit zero-target flag. Schema validation rejects malformed outputs, unknown EntityBank identifiers, and incomplete fields, then requests a corrected plan.

In Stage~3, the Refer-Planner performs closed-set selection over the identifiers and temporal segments supplied by EntityBank. It jointly checks identity, temporal, relational, cardinality, and exclusion constraints, so every prediction remains grounded in supported candidates. An explicitly absent target, or the absence of any candidate set satisfying all constraints, produces an empty selection. The selected Gaussian supports are rasterized only within their chosen temporal segments, and masks from multiple selected entities are combined frame by frame. This yields the final spatiotemporal segmentation while keeping the query interpretation, entity evidence, and referent decision inspectable.

\begin{figure*}[!t]
  \begin{minipage}[t]{\columnwidth}
    \centering
    \includegraphics[width=\linewidth]{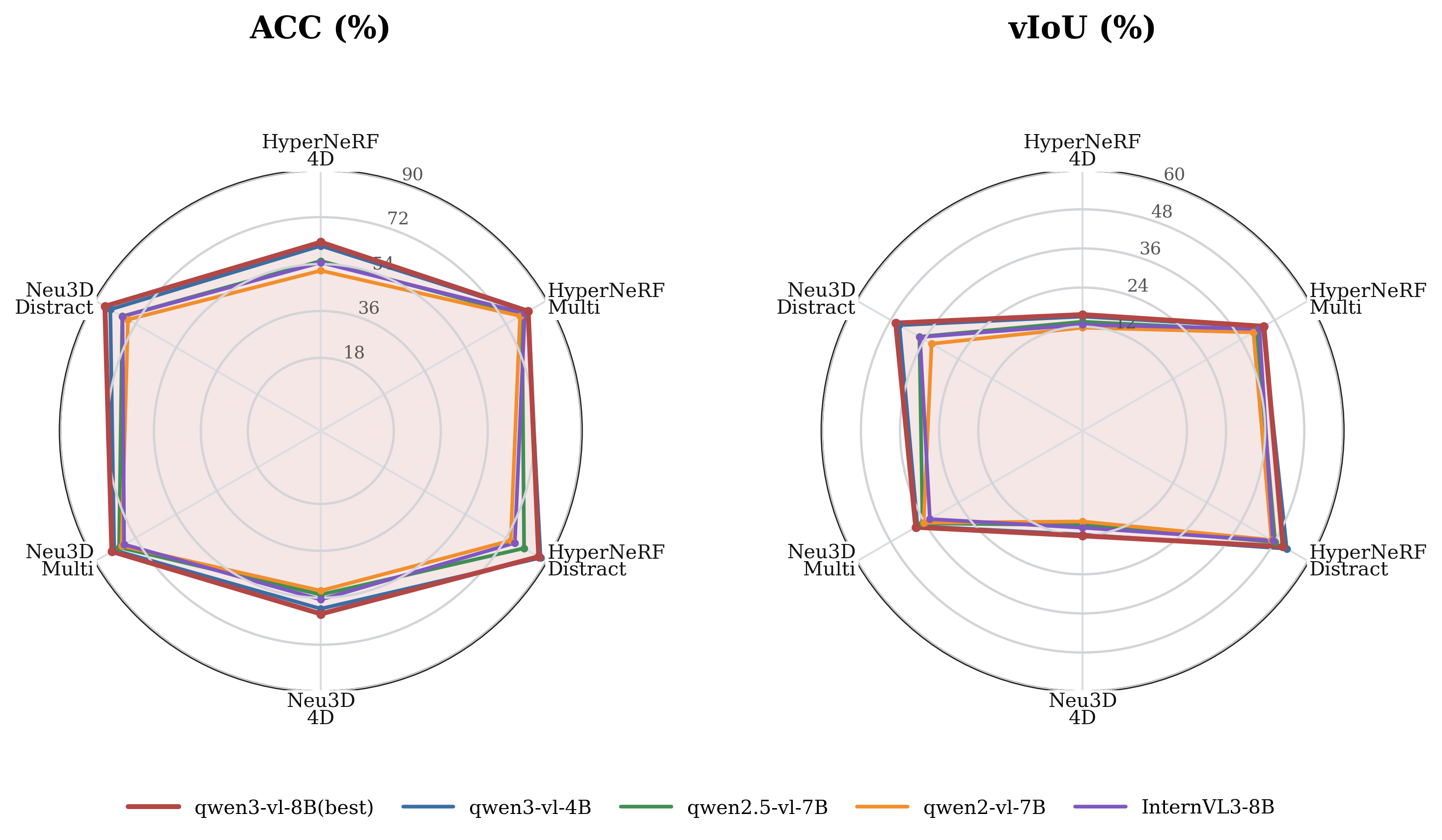}
    \captionof{figure}{Refer-Planner comparison by query category and scene source. Each panel reports six Acc or vIoU axes.}
    \Description{Radar comparison of five MLLM backbones across six referring metrics.}
    \label{fig:planner-sensitivity}
  \end{minipage}\hfill
  \begin{minipage}[t]{\columnwidth}
    \centering
    \includegraphics[width=\linewidth]{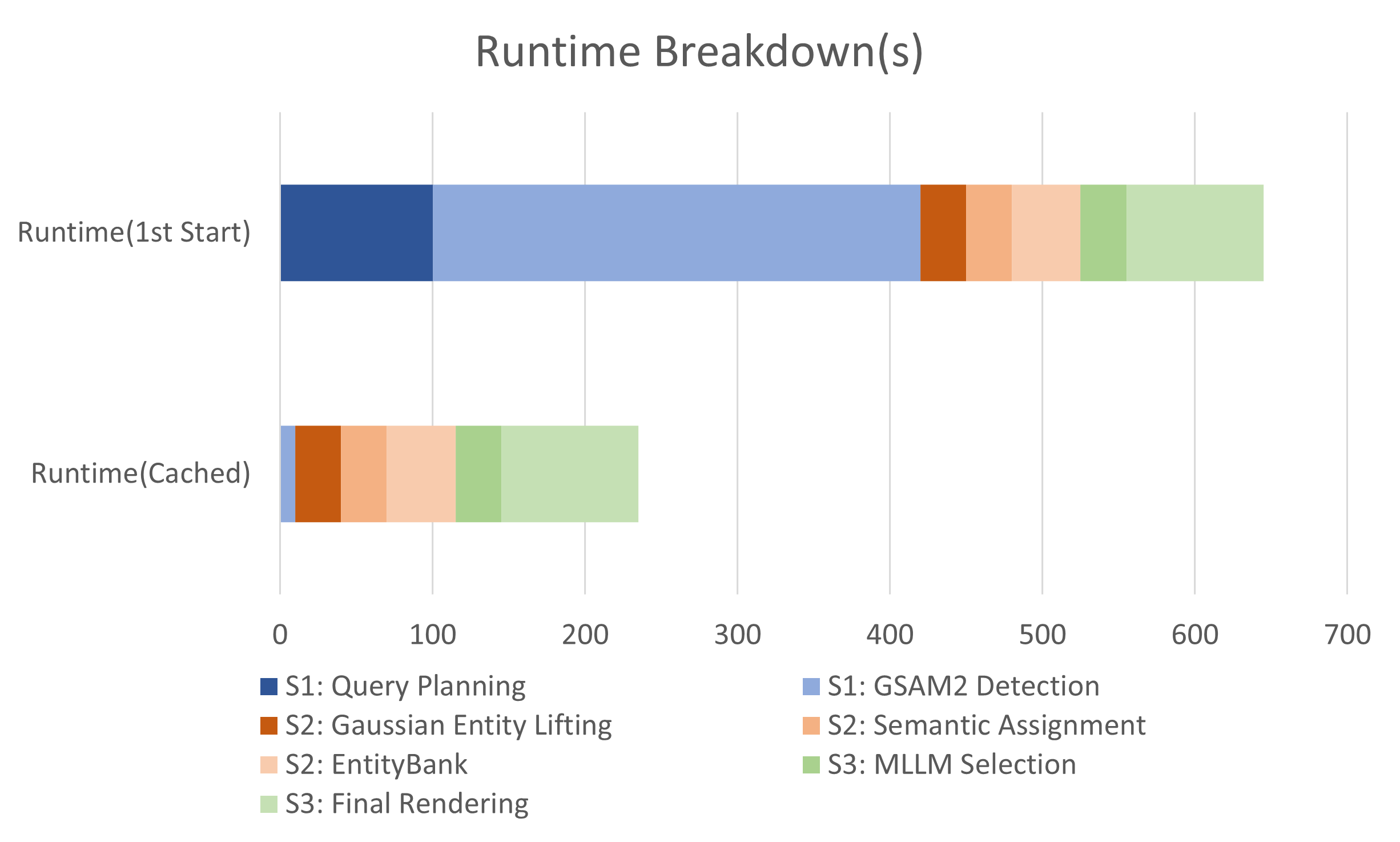}
    \captionof{figure}{Runtime for initial and cached queries, showing the benefit of EntityBank reuse. Colored segments report the cost of each pipeline component.}
    \Description{Runtime breakdown comparing the first query with subsequent queries that reuse cached entity evidence.}
    \label{fig:runtime}
  \end{minipage}
\end{figure*}

\section{Experiments}

\subsection{Experimental Setup}

We evaluate ReferGaussian on R4D-Bench-QA and the four-scene 4D LangSplat protocol~\cite{li20254dlangsplat}. All methods use the same scene inputs, query annotations, and evaluation protocols. We compare with 4D LangSplat and Segment then Splat~\cite{lu2025segmentthensplat}, the language-field and segmentation-first baselines. ReferGaussian uses Qwen3-VL-8B~\cite{bai2025qwen3} as the Refer-Planner and Grounded SAM 2~\cite{liu2023groundingdino,ren2024groundedsam,ravi2025sam2} to obtain phrase-conditioned mask tracks. The foundation models are used without additional training, and ReferGaussian performs no scene-specific semantic optimization.

Qwen3-VL-8B is the default planner. The planner analysis additionally evaluates Qwen3-VL-4B~\cite{bai2025qwen3}, Qwen2.5-VL-7B~\cite{bai2025qwen25vl}, Qwen2-VL-7B~\cite{wang2024qwen2vl}, and InternVL3-8B~\cite{zhu2025internvl3} under the same pipeline.

\subsection{Main Results}

\noindent\textbf{R4D-Bench-QA.}
Table~\ref{tab:main-r4dbench} shows that ReferGaussian reaches $76.5\%$ Acc and $34.4\%$ vIoU, surpassing both baselines on R4D-Bench-QA, which covers temporal, multi-target, and distractor queries.

\noindent\textbf{4D LangSplat protocol.}
Table~\ref{tab:4dlangsplat-protocol} reports Acc and vIoU for the four-scene protocol. ReferGaussian reaches $93.01\%$ average Acc and $62.95\%$ vIoU. Its Acc is $2.18$ points above 4D LangSplat, without per-scene semantic fitting.

\subsection{Qualitative Results}

Figure~\ref{fig:qual-results} shows coherent transition timing and clean exclusion results from entity-level temporal and set reasoning. Its columns show temporal-state and exclusion queries, while the rows compare RGB input, ground truth, ReferGaussian, Segment then Splat, and 4D LangSplat. In the liquid query, ReferGaussian follows the active filling interval; in the exclusion query, it preserves the requested objects while removing the hand that applies force. These examples illustrate temporal activation and set exclusion within the same entity-level inference process.

\balance
\subsection{Ablation and Analysis}

We analyze ReferGaussian from four perspectives: Refer-Planner choice, Gaussian entity construction, robustness to mask-evidence errors, and the efficiency gained through EntityBank reuse.

\noindent\textbf{Refer-Planner sensitivity.}
Figure~\ref{fig:planner-sensitivity} compares five MLLMs on 4D-varying, multi-target/reasoning, and zero-target/distractor queries. Acc and vIoU are reported separately on HyperNeRF and Neu3D, producing six source-category axes in each metric panel. Qwen3-VL-8B gives the strongest overall profile with stable trends across both scene sources.

\begin{table}[!t]
  \caption{Ablations of entity construction and mask evidence. Rows report vIoU for lifting variants and controlled mask perturbations.}
  \label{tab:functional-ablation}
  \centering
  \rfdtablefont
  \setlength{\tabcolsep}{3.0pt}
  \renewcommand{\arraystretch}{1.30}
  \begin{tabular*}{\columnwidth}{@{\extracolsep{\fill}}llc@{}}
    \toprule
    Group & Setting & vIoU (\%) \\
    \midrule
    Entity & Direct GSAM mask reference & 57.4 \\
    Entity & Framewise tracklet association & 38.2 \\
    Entity & One-pass Gaussian voting & 12.4 \\
    Entity & Iterative Gaussian lifting & 34.4 \\
    \midrule
    Mask & Original masks & 34.4 \\
    Mask & Erosion, radius $3/5$ & $33.8/32.1$ \\
    Mask & Dilation, radius $5$ & 32.6 \\
    Mask & Drop $10\%/25\%$ proposals & $33.9/32.5$ \\
    Mask & Merge/split errors & $19.4/22.7$ \\
    \bottomrule
  \end{tabular*}
\end{table}

Table~\ref{tab:functional-ablation} consolidates the functional-interface ablations. Direct GSAM masks and framewise tracklet association isolate the 2D evidence and temporal-association interfaces, while the Gaussian variants test how that evidence becomes persistent 4D support. Iterative lifting raises vIoU from $12.4$ to $34.4$ over one-pass voting by revisiting support with render-back feedback. Under mask perturbations, erosion, dilation, and proposal dropping retain $32.1$--$33.9$ vIoU, remaining within $2.3$ points of the original masks. Merge and split errors fall to $19.4$ and $22.7$ because they change entity identity and propagate to later constraints. The contrast separates tolerance to local mask noise from sensitivity to incorrect entity composition.

\noindent\textbf{Efficiency and reuse.} EntityBank reuse cuts runtime from about $600$ to $200$ seconds and local MLLM calls from six to three (Figure~\ref{fig:runtime}) using local inference. The initial query is dominated by mask detection and entity construction, while cached queries reuse tracks, visual evidence, and entity descriptions. The remaining work is concentrated in query planning, semantic assignment, selection, and final rendering.

\noindent\textbf{Failure modes.}
Remaining failures involve crowded contact events, ambiguous long queries, breakage, and severe occlusion. These cases combine rapid topology changes and uncertain identity boundaries, motivating stronger temporal association and calibrated uncertainty.

\section{Conclusion}
\label{sec:conclusion}

We formulated Referring 4D Gaussian Splatting (R4DGS), a new task for grounding complex natural-language expressions in dynamic Gaussian scenes, and introduced R4D-Bench-QA to evaluate temporal-state, multi-target, and zero-target referring. We further presented ReferGaussian, a training-free Gaussian-to-Entity framework that organizes dynamic Gaussian primitives into persistent, language-addressable entities and performs constraint-aware referent selection through EntityBank. Experiments on R4D-Bench-QA and the 4D LangSplat protocol demonstrate the effectiveness of entity-level reasoning for complex spatiotemporal grounding without scene-specific semantic optimization. Together, R4DGS, R4D-Bench-QA, and ReferGaussian establish a foundation for language-guided interaction and structured reasoning over dynamic 4D scene representations.

\clearpage
\nobalance
\bibliographystyle{ACM-Reference-Format}
\bibliography{ref,refergaussian_refs}

\end{document}